\documentclass[10pt, conference, letterpaper]{IEEEtran}
\IEEEoverridecommandlockouts
\usepackage{cite}
\usepackage{amsmath,amssymb,amsfonts}
\usepackage{graphicx}
\usepackage{tikz}
\usetikzlibrary{arrows.meta,positioning,fit,backgrounds,calc}

\usepackage{xcolor}

\usepackage{booktabs}

\usepackage{xcolor}
\usepackage{fvextra}
\definecolor{brA}{RGB}{200,40,40}
\definecolor{brB}{RGB}{30,120,200}
\definecolor{brC}{RGB}{120,80,180}
\newcommand{\lba}{\textcolor{brA}{\symbol{123}}}
\newcommand{\rba}{\textcolor{brA}{\symbol{125}}}
\newcommand{\lbb}{\textcolor{brB}{\symbol{123}}}
\newcommand{\rbb}{\textcolor{brB}{\symbol{125}}}
\newcommand{\lbc}{\textcolor{brC}{\symbol{123}}}
\newcommand{\rbc}{\textcolor{brC}{\symbol{125}}}
\newcommand{\lsb}{\textcolor{brB}{[}}
\newcommand{\rsb}{\textcolor{brB}{]}}
\DefineVerbatimEnvironment{JsonCode}{Verbatim}{
  commandchars=\\\<\>,
  fontsize=\footnotesize,
  fontfamily=tt,
  breaklines=true,
  framesep=3pt,
  xleftmargin=2pt,
  bgcolor=gray!8
}

\usepackage{textcomp}

\usepackage{url}

\usepackage{marvosym}
\usepackage{tabularx} 
\usepackage{algorithm}
\usepackage[noend]{algpseudocode}
\algrenewcommand\algorithmicrequire{\textbf{Input:}}
\algrenewcommand\algorithmicensure{\textbf{Output:}}

\usepackage{xspace}
\newcommand{\system}{MCPGuard\xspace}
\newcommand{\ie}{{\em i.e., \xspace}}
\newcommand{\etc}{{\em etc\xspace}}
\newcommand{\eg}{{\em e.g., \xspace}}

\newcommand{\squishlist}{
\begin{list}{$\bullet$}{
  \setlength{\itemsep}{0pt}
  \setlength{\parsep}{3pt}
  \setlength{\topsep}{3pt}
  \setlength{\partopsep}{0pt}
  \setlength{\leftmargin}{3.5mm}
  \setlength{\labelwidth}{1em}
  \setlength{\labelsep}{0.5em}}}
\newcommand{\squishend}{\end{list}}

\begin{document}

\title{\system: Finding Conformance Gaps\\ in MCP SDKs via Dual-Track Analysis\\ and Dynamic Adjudication
\thanks{Preprint. This manuscript is currently under review.
}}


\author{
\IEEEauthorblockN{
Xikai Sun\IEEEauthorrefmark{1},
Kebin Liu\IEEEauthorrefmark{1},
Fan Dang\IEEEauthorrefmark{2}, 
Yawen Zheng\IEEEauthorrefmark{1},
Zihao Yang\IEEEauthorrefmark{3}, 
Yunhao Liu\IEEEauthorrefmark{1}\textsuperscript{\Letter}
}

\IEEEauthorblockA{
\IEEEauthorrefmark{1}Tsinghua University     
\IEEEauthorrefmark{2}Beijing Jiaotong University
\IEEEauthorrefmark{3}Yanshan University
$^\textrm{\Letter}$ Corresponding Author
}

}

\maketitle

\begin{abstract}
The Model Context Protocol (MCP) connects large language model agents to external tools and services through official SDKs that implement message validation and session logic. Specification constraints omitted by these SDKs may propagate to downstream applications, highlighting the importance of SDK-level conformance testing. Yet existing MCP conformance tools mainly exercise running endpoints and reveal little about whether SDK models enforce specification requirements. We present MCPGuard, an automated LLM-based framework for detecting specification-to-SDK conformance gaps. Its structural track aligns JSON Schema constraints with normalized SDK models and produces constraint-specific evidence. Its behavioral track extracts normative prose rules, retrieves bounded function-level slices, and audits them with LLMs for candidate omissions. An evidence-aware dynamic adjudicator then inspects complete source context and builds PoCs to validate candidates. A shared kernel with language-specific adapters enables analysis across SDKs. On the official Python 1.26.0 and TypeScript 1.29.0 SDKs, MCPGuard confirms 13 and 19 findings, respectively, none of which are exposed by two black-box baselines. Tests on open MCP servers further show that these gaps can systematically propagate downstream. By making SDK conformance of specification constraints directly testable, MCPGuard complements conventional endpoint-oriented conformance suites.

\end{abstract}


\section{Introduction}
The Model Context Protocol (MCP) has become a widely adopted standard for connecting large language model (LLM) agents to external tools, data, and services~\cite{mcp-launch,mcp-aaif}. Since its release in November 2024, it has grown from a single-vendor interface into a broad ecosystem of agents and servers~\cite{mcp-first-glance}. Many MCP servers rely on a set of official SDKs rather than implementing the protocol from scratch. The Python and TypeScript SDKs record over 97 million downloads per month~\cite{mcp-aaif}.
These SDKs therefore form a shared layer between the specification and downstream deployments.
Consequently, a constraint omitted by an SDK is not limited to one server implementation. It can be inherited by the clients and servers built on that SDK unless they add independent validation.

This shared dependency makes SDK conformance important. 
The MCP specification has undergone several dated revisions and introduced features such as elicitation, tasks, and new transport rules~\cite{mcp-spec}. Reference SDKs may not immediately reflect every specification change. This creates persistent drift between the specification and its SDK implementations~\cite{compatibility-cost,agentrfc}. Such drift may also arise from incomplete validation, implementation choices, or limitations in SDK modeling. Specification-to-implementation drift is thus a standing risk for a protocol whose libraries evolve alongside its contract. 
This drift is defined as the \textbf{conformance gap}, \ie the constraint declared by the specification but not enforced by the SDK.

Conformance gaps can affect SDKs' correctness and interoperability and weaken validation across trust boundaries. An MCP host allows an LLM agent to exchange messages with third-party servers that the host does not control~\cite{mcp-spec,mcp-threat}. The SDK parses these messages and is expected to enforce the constraints defined by the protocol's specification. Consider a field that the specification defines as a URI. If the SDK validates it only as an unrestricted string, malformed values may enter downstream code that renders, stores, or dereferences the field.  The missing validation violates the protocol contract and may create security-relevant behavior. Despite the central role of official SDKs, their enforcement of specification constraints has not been systematically studied.

Although there has been some research on identifying defects, existing approaches suffer from one or more of the following limitations.
\begin{itemize}
	\item Hand-written conformance tests are difficult to maintain. Manual auditing requires experts to identify relevant requirements, locate their implementations, and construct violating inputs. The process must be repeated as the specification and SDKs evolve. This limits both coverage and scalability~\cite{rfcscan,protocolguard,matterguard}. 
	\item Existing black-box testing tools observe protocol behavior but not SDK data models~\cite{mcp-conformance,yaw-compliance}. They send requests to a running server and inspect its responses. They cannot directly determine whether typed SDK models enforce field-level constraints such as numeric ranges and required fields. They may also miss a gap when their own client SDK prevents them from constructing the invalid input needed to expose it.
	\item Black-box observations provide limited attribution. A gap observed by probing a deployed server cannot be traced to the SDK versus the server's own code. Therefore, it neither localizes the fix nor establishes that the gap is systematic. 
\end{itemize}
Due to these limitations, prior approaches are not suitable for our purpose.  These limitations call for an automated method that compares the specification directly with the SDK and verifies each detected inconsistency against the real implementation.

We present MCPGuard, an automated method for detecting specification-to-SDK conformance gaps for MCP.
It follows two tracks matched to the specification's two representations. The structural track reflects over typed SDK models and deterministically compares them with the machine-readable schema. The behavioral track extracts normative behavioral rules from spec documents, selects rule-relevant functions using our lexical slicer, and asks an LLM to identify candidate omissions. 
A final adjudication stage checks the complete implementation context and builds PoCs to verify the authenticity of gaps. By sharing a kernel and using language-specific adapters, MCPGuard can test multiple SDKs.

Applied to the two official SDKs, MCPGuard confirms 13 findings in Python and 19 in TypeScript. The structural track contributes 11 and 16 findings, all reproduced by generated PoCs; full-context review retains two and three behavioral findings. We reported the findings privately to the maintainers. On the same SDK versions, the official conformance suite passes all 42 applicable checks, and a third-party suite assigns grades of 96 and 94, yet neither baseline exposes any MCPGuard finding. A measurement of 32 authorized public servers further shows that two shared URI-format gaps are reachable: all 32 endpoints that complete the handshake accept both malformed metadata fields.

Our main contributions are as follows:
\begin{itemize}
    \item We present MCPGuard, a specification-to-SDK framework for MCP conformance testing. Given the increasing importance and widespread adoption of MCP, MCPGuard is crucial for ensuring the functionality and reliability of MCP applications.
    \item We propose a novel testing mechanism, \ie dual-track analysis. The structural track aligns machine-readable constraints with SDK validators, and the behavioral track uses rule-guided code slicing and LLM reasoning to identify behavioral candidates. Dynamic adjudication checks the authenticity of those candidates.
	\item MCPGuard enables testing of different SDKs by sharing a kernel and language-specific adapters. A cross-language evaluation on the official Python and TypeScript SDKs yields 27 structural and 5 behavioral gap findings. Detailed experiments demonstrate the systematic downstream propagation of those gaps.
\end{itemize}

\section{Related Work}

\textbf{MCP security.}
Existing MCP security research has primarily studied attacks that exploit the semantic authority granted to tools~\cite{mcp-security-bench,mcp-attack-vectors,mcpsecbench}. MCPTox evaluates tool-poisoning attacks embedded in tool metadata across real-world MCP servers and agent configurations~\cite{mcptox}. Another work develops threat models for MCP hosts, clients, servers, and authorization components, identifying prompt injection and tool poisoning as central risks~\cite{mcp-threat}. This line of work mainly asks whether protocol-valid content can manipulate an agent after it has been accepted. MCPGuard examines an earlier enforcement boundary: whether the SDK accepts a message or protocol action that already violates the structural or behavioral constraint declared by the MCP specification.
Recent MCP-specific work moves closer to SDK testing. {Compatibility at a Cost} constructs a language-independent representation of multiple MCP SDKs and combines auditable static analysis with LLM-guided semantic reasoning to identify clause-compliance inconsistencies and compatibility-abusing attacks~\cite{compatibility-cost}. AgentRFC extracts agent-protocol clauses into a typed protocol IR, compiles security principles into TLA+ invariants, and replays model-checking counterexamples against live SDKs~\cite{agentrfc}. These systems study static security invariants or exploitability. 
MCPGuard instead uses static data analysis and dynamic session verification to conduct in-depth testing of the consistency of the MCP SDK.

\textbf{Schema-driven and conformance testing.}
Schema-driven API testing uses a machine-readable contract to generate inputs and evaluate deployed services~\cite{miner,nautilus,arat-rl}. For example, QuickREST derives property-based generators and validation properties from OpenAPI descriptions, while Schemathesis derives structure and semantics fuzzers from OpenAPI or GraphQL schemas~\cite{quickrest,schemathesis}. RESTler infers producer--consumer dependencies from API specifications to construct stateful request sequences, and MOREST refines a service-dependency model using execution feedback~\cite{restler,morest}. These methods use the schema as an external test generator or oracle and observe endpoint execution. They do not directly determine whether a protocol constraint was lost when translated into an internal SDK model.
MCP's official conformance framework also drives client and server scenarios, captures protocol interactions, and checks observable behavior. Similarly, the Yaw Labs suite performs capability-aware compliance checks against MCP server endpoints~\cite{mcp-conformance,yaw-compliance}. These frameworks provide valuable tools, but they do not explore how MCP-specific constraints are represented by SDK models. 
MCPGuard complements those conventional suites by making the SDK conformance of specification constraints directly testable.


\textbf{LLM-assisted implementation analysis.} Beyond synthesizing tests from specifications, another line of work uses LLMs to relate normative rules directly to implementation code~\cite{parcleanse,llift,llmxcpg,fuzz4all}. For example,
RFCScan builds hierarchical semantic indexes over RFC documents and protocol source code, and uses demand-driven retrieval to identify functional inconsistencies~\cite{rfcscan}. ProtocolGuard combines specification rule extraction, LLM-guided program slicing, assertion instrumentation, and fuzzing-based confirmation for conformance testing of conventional protocol implementations~\cite{protocolguard}. LLM-guided fuzzers such as ChatAFL instead recover message grammars and state-transition guidance to improve dynamic protocol exploration~\cite{chatafl}. At the SDK layer, MatterGuard demonstrates specification-guided testing for conformance defects in Matter SDKs~\cite{matterguard}. 
Compared to these works, MCPGuard adopts a narrower oracle. Machine-readable constraints are compared deterministically with SDK validators, while LLM reasoning is restricted to judging behavioral-rule candidates and assisting dynamic adjudication.
\begin{figure}[!t]
	\centering
	\includegraphics[width=0.95\columnwidth]{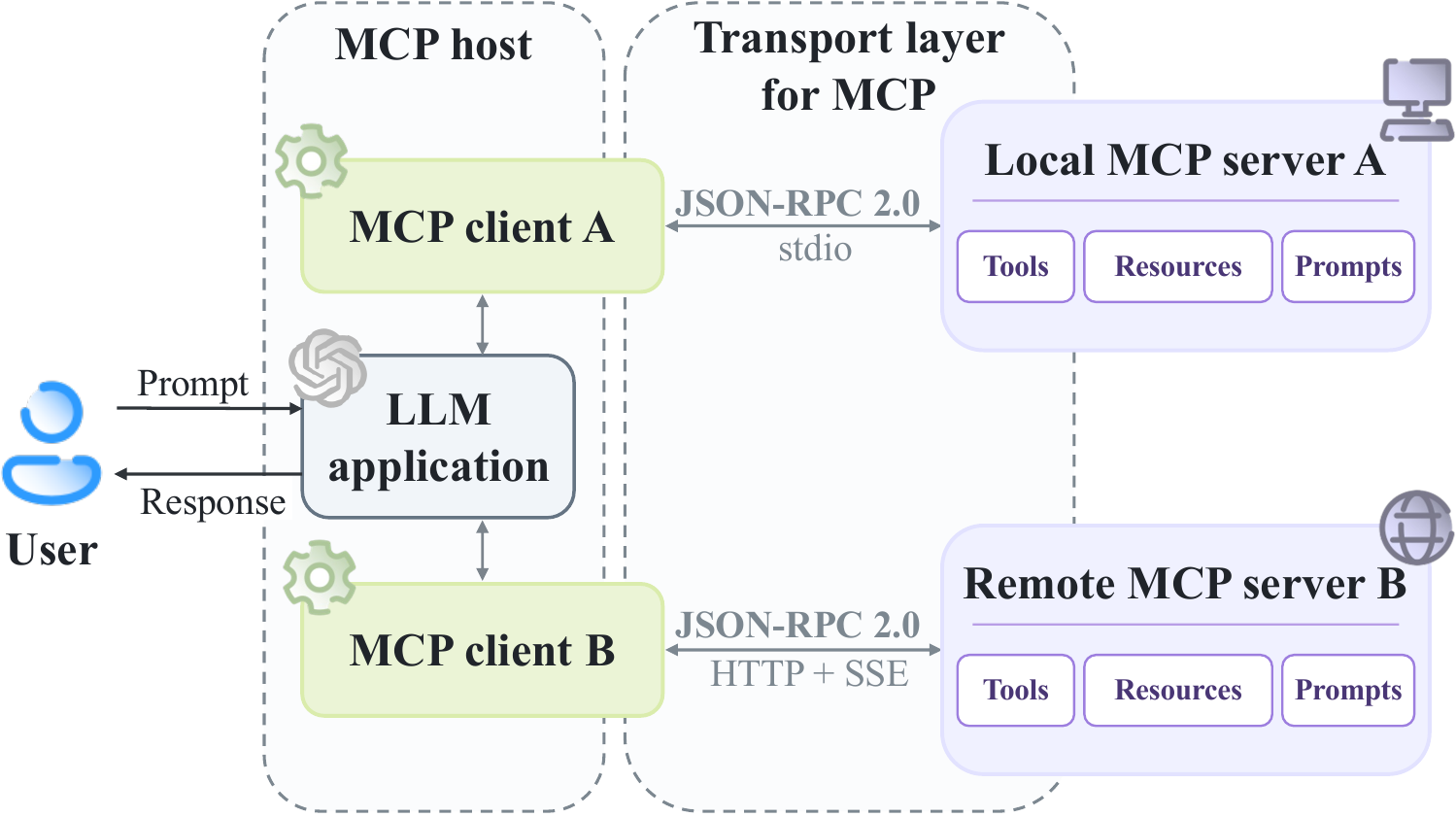}
	\caption{MCP host--client--server architecture.}
	\label{fig:mcp-mechanism}
\end{figure}

\section{Background and Problem Model}
\label{sec:background}

MCP conformance spans message structure, session semantics, and the SDK code that mediates both. We introduce the protocol objects used later, position official SDKs on the message path, and formalize two classes of conformance gap.

\subsection{The Model Context Protocol}
\label{subsec:mcp}
MCP follows a three-party {client--host--server} architecture~\cite{mcp-spec}, as shown in Fig.~\ref{fig:mcp-mechanism}. A host is the LLM application and creates one client for each configured server; each client maintains an isolated, stateful session with that server. Servers may be local processes or remote services and expose \texttt{tools} for actions, \texttt{resources} for contextual data, and \texttt{prompts} for reusable templates. Clients can offer roots, sampling, and elicitation. For example, a client discovers tools with \texttt{tools/list}; after the model selects one, the host routes \texttt{tools/call} through the corresponding client and may add the returned content to the model context.
MCP encodes requests, responses, and notifications as JSON-RPC 2.0~\cite{jsonrpc20}. A request has an \texttt{id}, a \texttt{method}, and optional \texttt{params}; its response repeats the \texttt{id} with a \texttt{result} or \texttt{error}. A notification has no request identifier. The same message objects travel over \texttt{stdio} for a local server or Streamable HTTP for a remote endpoint, optionally using Server-Sent Events for server-to-client traffic~\cite{mcp-spec}.

A session starts with \texttt{initialize}, in which the client sends its protocol version, identity, and capabilities. The server returns its selected version, identity, and capabilities; the client then sends \texttt{notifications/initialized}. Subsequent operations must respect that negotiation. For instance, \texttt{sampling/createMessage} requires client sampling support, and elicitation separately negotiates \texttt{form} and \texttt{url} modes. A message can therefore be structurally valid yet illegal in its session context~\cite{mcp-spec}.

\subsection{MCP Specification and SDKs}

The MCP specification presents its contract in two forms, including machine-readable schema and prose documents.
The specification defines its TypeScript schema as the source of truth and publishes an automatically generated JSON Schema to expose field declarations to automated tooling~\cite{mcp-spec}.
The documents define lifecycle ordering, capability conditions, and responsibilities that are not isolated field schemas.

In the SDK versions evaluated in our work, the official Python and TypeScript SDKs translate these sources into Pydantic models and Zod models, respectively~\cite{mcp-python,mcp-typescript}.
On a standard inbound path, an SDK decodes a JSON-RPC object, parses it into a typed message, applies session checks, and dispatches it to an application handler. Pydantic or Zod primarily enforces requiredness, enumerations, ranges, lengths, and formats. Client, server, and session logic defined in SDKs enforces initialization and capability rules. 

\subsection{Conformance Gap Model}
\label{subsec:gap}

The conformance constraints of MCP include structural and behavioral levels.
Let $\mathcal{T}$ be the message types,
and let $L(t)$ denote the schema locations within type $t\in\mathcal{T}$, including the type root, fields, object members, and nested objects.
The structural constraints $C_s$, extracted from the JSON Schema~\cite{jsonschema-validation}, contain constraints $c=(t,\ell,\kappa,e)$, where $\ell\in L(t)$ identifies the constrained schema location, $\kappa$ is the constraint kind, and $e$ records its declared expectation. We consider
\begin{equation}
    \mathcal{K}_{s}=
    \{\textsf{required},\textsf{const},\textsf{enum},
    \textsf{range},\textsf{format},\textsf{len}\},
    \label{eq:kinds}
\end{equation}
such that $\kappa\in\mathcal{K}_s$ for every $c\in C_s$.
The behavioral constraints $C_b$ contain rules expressed in the documents, typically marked by BCP~14 keywords such as \textsc{MUST} and \textsc{MUST NOT}~\cite{rfc2119,rfc8174}. If $V_I\subseteq C_s\cup C_b$ denotes the constraints enforced by SDK implementation $I$, then the conformance gap is as follows:
\begin{equation}
	\mathit{Gap}(I)=
	\underbrace{C_s\setminus V_I}_{\mathit{Gap}_s(I)}
	\cup
	\underbrace{C_b\setminus V_I}_{\mathit{Gap}_b(I)}.
	\label{eq:gap}
\end{equation}

A structural gap $c\in\mathit{Gap}_s(I)$ means that implementation $I$ fails to enforce a schema constraint and consequently accepts at least one message that the schema rejects. Such a discrepancy may affect either message structure or field semantics: a required property may be absent, and a constant or enumeration value may be illegal. For example, in the server-to-client \texttt{sampling/createMessage} request ($t$), \texttt{ModelPreferences.costPriority} ($\ell$) is constrained to a number in $[0,1]$(\ie $e=[0,1]$, and $\kappa=\textsf{range}$). An SDK that accepts \texttt{1001.0} preserves the expected JSON type but violates the declared range~\cite{mcp-spec}.

\begin{figure*}[tbp]
	\centering
	\includegraphics[width=2\columnwidth]{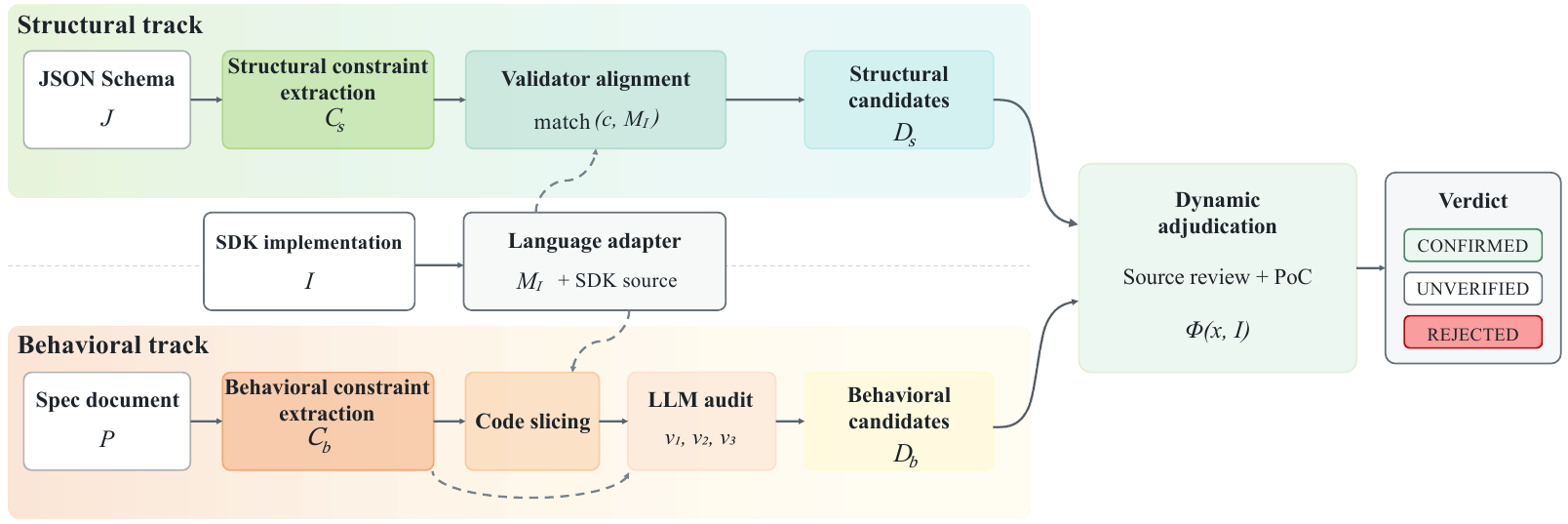}
	\caption{Overview of MCPGuard. It uses separate candidate-generation tracks for structural and behavioral constraints, and conducts independent source review and PoC-based runtime adjudication to determine whether candidate gaps are genuine.}
	\label{fig:overview}
\end{figure*}

By contrast, a behavioral gap $r\in\mathit{Gap}_b(I)$ permits a protocol trace forbidden by the spec document, even when every individual message is schema-valid.
For example, after a client advertises support for \texttt{elicitation.form} but not \texttt{elicitation.url}, a server must not issue a URL-mode \texttt{elicitation/create} request~\cite{mcp-spec}. Confirming this violation cannot rely on the request payload alone. Its witness must reconstruct the relevant trace, including the participant roles, completion of initialization, advertised capabilities, and the subsequent handling of the prohibited request.


\section{Method: MCPGuard}
\label{sec:method}

MCPGuard turns the conformance model in Eq.~\eqref{eq:gap} into two concrete questions. 
Formally, the conformance model focuses on two issues: first, does an SDK validator preserve the structural constraints declared by the JSON Schema? Second, does the SDK control logic enforce the behavioral constraints stated in the spec documents?
As Fig.~\ref{fig:overview} shows, MCPGuard answers the first question with deterministic constraint matching and uses an LLM to screen code for the second. Both tracks then pass their candidates to the final evidence-based dynamic adjudication step.
This separation follows the form of the specification. A range or requiredness constraint has a direct representation on both sides and can be compared mechanically. In contrast, a capability or lifecycle rule describes several messages and program states, so it must be interpreted together with the relevant control flow. 
Keeping these tasks separate avoids using probabilistic reasoning where an exact comparison is available, while the shared adjudication step prevents either a static mismatch or an LLM judgment from being reported directly as a finding.

\subsection{Building Comparable Views of the specification and SDK}
\label{subsec:extract-spec}

MCPGuard first converts the specification and the SDK into records that can be aligned. On the specification side, MUPGuard uses an extractor to visit each named object under the JSON Schema \texttt{\$defs} block and records the supported constraints on its direct properties. This instantiates the general tuple $c=(t,\ell,\kappa,e)$ from \S\ref{subsec:gap}; $e$ stores the expected presence, literal, set, interval, format, or length. 
In implementation, the schema extractor records \texttt{REQUIRED} for properties listed in the parent object's \texttt{required} array and \texttt{OPTIONAL} for the remaining properties. We normalize both cases as $c=(t,\ell,\textsf{required},e)$, where $e$ is a Boolean expectation. For example, the relevant schema fragment  of $t_{\mathrm{call}}=\texttt{CallToolRequestParams}$ is

\begin{JsonCode}
\lba
  "properties": \lbb
    "name": \lbc"type": "string"\rbc,
    "arguments": \lbc"type": "object"\rbc
  \rbb,
  "required": \lsb"name"\rsb
\rba
\end{JsonCode}
The extractor emits a \texttt{REQUIRED} record for \texttt{name} and an \texttt{OPTIONAL} record for \texttt{arguments}. We normalize these two implementation records using a single constraint kind with a Boolean expectation:
\begin{equation}
    \begin{aligned}
        c_{\mathrm{name}}
        &=(t_{\mathrm{call}},\texttt{name},
        \textsf{required},\mathsf{true}),\\
        c_{\mathrm{args}}
        &=(t_{\mathrm{call}},\texttt{arguments},
        \textsf{required},\mathsf{false}),
    \end{aligned}
    \label{eq:required-normalization}
\end{equation}
where $e=\mathsf{true}$ denotes a required property, whereas
$e=\mathsf{false}$ denotes an optional property.

In parallel, MCPGuard scans the Markdown spec documents for emphasized BCP~14 keywords and stores the complete sentence in an audit pool $C_b$. The mandatory, applicable rules (\eg rules with the keyword \textsc{MUST}) correspond to the behavioral constraints $C_b$ defined in \S\ref{subsec:gap}. Each rule is annotated with its source location and its JSON-RPC method. These annotations can help locate SDK implementation code.

On the SDK side, MCPGuard introduces adapters for language-specific SDKs to produce a normalized view $M_I$. For every extracted property, this view records whether the corresponding SDK field exists and which input constraints it enforces, including requiredness, literals, enumerations, numeric bounds, string bounds, validator patterns, and aliases. 
For example, the schema constrains \texttt{ModelPreferences.costPriority} to the interval $[0,1]$ when the field is present. The Python adapter locates both the \texttt{ModelPreferences} model and its \texttt{costPriority} field in the Python SDK. It records the field as optional with the annotation \texttt{float\,|\,None}, but finds neither a lower-bound nor an upper-bound validator. In the normalized view, \[M_I(\texttt{ModelPreferences},\texttt{costPriority}). \textsf{range}=\bot,\] where $\bot$ denotes missing enforcement. Thus, the schema field and SDK field are successfully aligned, while their range constraints are not.
At the type level, MCPGuard also checks whether each object named in JSON Schema has an aligned SDK model. This separate check captures whole-type modeling gaps rather than reducing all structural candidates to individual fields.




\subsection{Structural Track: Matching Structural Constraints}
\label{subsec:detect}

For each extracted $c\in C_s$, the structural track checks whether $M_I$ enforces the same expectation at the corresponding property. 
It first checks the coverage of each schema object in $M_I$. 
Let $a_I(t)$ return the SDK model aligned with schema type $t\in \mathcal{T}$; $a_I(t)=\bot$ indicates that no aligned model exists. The type-level candidates $D_s^{\mathrm{type}}(I)$ are as follows:
\begin{equation}
    D_s^{\mathrm{type}}(I)=
    \{t\in\mathcal{T}
    \mid a_I(t)=\bot\}.
    \label{eq:ds-type}
\end{equation}

For every type with an aligned SDK model, MCPGuard then compares its property constraints.
According to $\kappa$, the comparison covers
requiredness, constant values, enumeration sets, the presence of
declared numeric or string bounds, and format-validation signals.
The property-level candidates $D_s^{\mathrm{prop}}(I)$ are as follows:
\begin{equation}
\begin{aligned}
    D_s^{\mathrm{prop}}(I)=
    \{c\in C_s
    \mid {}&
    a_I(t)\neq\bot \\
    &{}\land
    \neg\mathrm{match}
    (c,M_I(t,\ell))\\
    &{}\land\neg\rho(c)\},
\end{aligned}
\label{eq:ds-prop}
\end{equation}
where $\mathrm{match}$ implements the comparison for constraint kind $\kappa$. 
The function $\rho$ summarizes explicit representation rules in the implementation.
For example, it skips \texttt{jsonrpc} and \texttt{id} when an outer model carries the JSON-RPC envelope, accepts a defaulted \texttt{Literal} as a method discriminant, and aligns known aliases and union/mixin types. These explicit rules prevent representation-equivalent designs of SDKs from being reported as missing validation.

Finally, the complete structural candidates are as follows:
\begin{equation}
    D_s(I)=
    D_s^{\mathrm{type}}(I)
    \cup
    D_s^{\mathrm{prop}}(I),
    \label{eq:ds}
\end{equation}

\textbf{Structural track example.} The schema extractor emits $c=(\texttt{ModelPreferences},\texttt{costPriority},\textsf{range},[0,1])$. The Python adapter observes \texttt{float\,|\,None} with no \texttt{Ge}/\texttt{Le} metadata, so $\mathrm{match}$ is false and no suppression rule applies. Equation~\eqref{eq:ds} therefore emits $c$. During adjudication in \S~\ref{subsec:adjudication}, the run-time verifier supplies \texttt{1001.0}; the real Pydantic constructor accepts it. 
The emitted regression PoC uses \texttt{999.0}, an equivalent out-of-range witness. 


\subsection{Behavioral Track: Auditing Behavioral Constraints}

A behavioral rule is expressed as a natural-language statement and cannot be matched to one validator.
MCPGuard therefore introduces LLMs in the behavioral track.
Its SDK implementation may be distributed across protocol dispatchers, capability guards, and session handlers. Supplying the entire SDK to the LLM introduces substantial irrelevant context, while extracting only the lines surrounding a keyword can omit the guard conditions and error paths needed to judge conformance. 
We therefore design a rule-guided, function-level slicer that retrieves complete candidate functions under a bounded context budget.

\textbf{Language-independent retrieval.}
Given a behavioral rule $r$, MCPGuard first derives the search-token set $\mathrm{tok}(r)$ from its associated JSON-RPC method, back-ticked identifiers, camel- and pascal-case identifiers, and MCP behavior terms. 
Let $\mathcal{F}_I$ be the ordered source-file list of SDK $I$, and let $\mathrm{Func}(p)$ denote the functions from source file $p$. MCPGuard constructs the slice $q_I(r)$ as follows:
\begin{equation}
\begin{aligned}
q_I(r)=\operatorname{Prefix}_{B}\Big(
\operatorname{Dedup}\big[
\!\!\bigoplus_{p\in\mathcal{F}_I}
\{\,\langle p,f\rangle \mid\;&
f\in\mathrm{Func}_{I}(p),\\
\exists\tau\in\mathrm{tok}(r):
\tau\sqsubseteq f\,\}
\big]\Big),
\end{aligned}
\label{eq:rule-guided-slice}
\end{equation}
where $\tau\sqsubseteq f$ denotes a case-insensitive occurrence, $\bigoplus$ preserves the core-first file order, $\operatorname{Dedup}$ removes repeated $\langle p,f\rangle$ pairs, and $\operatorname{Prefix}_{B}$ limits the resulting context to $B=4{,}000$ characters. Each entry retained in $q_I(r)$ contains the source path, function name, and recovered complete function body.

\textbf{Language-specific boundary recovery.}
Different SDKs require specific mechanisms for recovering reliable function boundaries. For example, for the Python SDK, MCPGuard parses each source file into an abstract syntax tree and visits \texttt{FunctionDef} and \texttt{AsyncFunctionDef} nodes. The node's \texttt{lineno} and \texttt{end\_lineno} identify the complete function body, including nested conditions and exception paths. For the TypeScript SDK, MCPGuard uses regular-expression patterns to recognize function declarations, class methods, and arrow functions. Starting from a matched declaration, it tracks the brace depth until the corresponding closing brace.
Although the two backends use different syntax mechanisms, they expose a similar function-level slice to the downstream LLM.


\textbf{LLM audit.} The LLM receives one rule and its candidate slice and decides whether the code appears to omit the required guard. We query each rule three times and retain a behavioral candidate $r\in D_b(I)$ only when a majority reports a violation with agreement of at least $0.6$. Voting reduces sensitivity to one decoding path, but it does not establish correctness~\cite{self-consistency}. If no relevant function can be located, the rule is marked \textsc{skipped-no-code}; absence of code evidence is never interpreted as conformance.

\textbf{Behavioral track example.}
Consider the rule ``Servers \textsc{MUST NOT} send \underline{elicitation} requests with modes that are not supported by the client.'' The tokens include
\texttt{elicitation/create}, \texttt{create}, and \texttt{elicit}.
MCPGuard locates \texttt{assertCapabilityForMethod} and recovers its complete body. The resulting slice exposes that the function checks only whether \texttt{clientCapabilities.elicitation} exists, but does not compare the requested mode with the advertised \texttt{form} or \texttt{url} modes. 



\subsection{Dynamic Adjudication}
\label{subsec:adjudication}

The two tracks meet at an adjudicator $\Phi$ that assigns \textsc{confirmed}, \textsc{rejected}, or \textsc{unverified}. A candidate is \textsc{confirmed} only when independent evidence demonstrates the conformance gap. It is \textsc{rejected} when broader source context or execution shows that the constraint is enforced, and \textsc{unverified} when the available evidence cannot decide. 

Structural candidates are checked with kind-specific witness templates. MCPGuard first populates the required fields with simple seed values, then removes a required member or supplies an out-of-range number, over-length string, illegal enumeration value, malformed URI, or invalid Base64 string. 
Then, MCPGuard invokes the actual Pydantic constructor or Zod \texttt{safeParse} to check the validity of the constructed inputs. 
Acceptance confirms the conformance gap; rejection refutes the candidate. 

For behavioral candidates, the adjudicator $\Phi$ first reopens the complete function and its relevant callers and callees to verify the responsible peer, SDK ownership, and applicable protocol state. When a violating trace can be constructed, MCPGuard encodes it as a minimal executable Proof-of-Concept (PoC) using an LLM and runs it against the SDK in an isolated local environment.
Depending on the message direction, MCPGuard either creates a fresh client--server pair over the SDK's in-memory transport or launches the target server as a fresh stdio subprocess and drives it with an SDK-independent raw JSON-RPC peer. The latter prevents a validator in the testing-side SDK from blocking the malformed message before it reaches the target. Each PoC establishes the required initialization state and advertised capabilities, executes the minimal message sequence ending in the offending action, and records the resulting JSON-RPC response, error, callback entry, process failure, or timeout.
The session or subprocess is terminated after each run to isolate its state from subsequent probes.
A candidate is dynamically confirmed only when the real protocol trace accepts the disallowed action or reaches a callback that the specification requires the SDK to block. 

A rejected pre-initialization candidate illustrates why adjudication is necessary. 
Its local code slice appears to contain no initialization-state check, causing the LLM audit to flag a potential violation. However, an end-to-end in-memory session probe shows that an outer session layer has rejected the request and returned the JSON-RPC error \texttt{-32602}~\cite{jsonrpc20}.


Finally, the reported finding set is as follows:
\begin{equation}
	\widehat{\mathit{Gap}}(I)=
	\{x\in D_s(I)\cup D_b(I)|\Phi(x)=\textsc{confirmed}\}.
	\label{eq:confirmed-gap}
\end{equation}
The MCPGuard pseudocode is shown in Algorithm~\ref{alg:mcpguard}.

%
%
%

\begin{algorithm}[t]
	\caption{MCPGuard}
	\label{alg:mcpguard}
	\begin{algorithmic}[1]
		\small
		
		\Statex \textbf{Input:} JSON Schema $J$, prose specification documents $P$, and SDK implementation $I$.
		
		\Statex \textbf{Output:} Confirmed conformance-gap set $\widehat{\mathit{Gap}}(I)$.
		
		\State $C_s\leftarrow\textsc{ExtractStruct}(J)$
		
		\State $\mathcal{C}_b\leftarrow\textsc{ExtractRules}(P)$
		
		\State $M_I\leftarrow\textsc{AdaptModels}(I)$
		
		\State $D_s\leftarrow\textsc{MissingTypes}(J,M_I)$
		
		\State \textbf{for each} $c=(t,\ell,\kappa,e)\in C_s$ \textbf{do}
		
		\State \quad \textbf{if} $\neg\mathrm{match}(c,M_I(t,\ell))\land\neg\rho(c)$ \textbf{then} $D_s\leftarrow D_s\cup\{c\}$
		
		\State $D_b\leftarrow\emptyset$
		
		\State \textbf{for each} $r\in\textsc{Prioritize}(\mathcal{C}_b)$ \textbf{do}
		
		\State \quad $q\leftarrow\textsc{Slice}(I,\mathrm{tok}(r),4000)$
		
		\State \quad \textbf{if} $q=\emptyset$ \textbf{then record \textsc{skipped-no-code}; continue}
		
		\State \quad $v_1,v_2,v_3\leftarrow\textsc{LLMAudit}(r,q)$
		
		\State \quad \textbf{if} $\mathrm{majority}(v_i)=\textsf{violation}\land\mathrm{agree}(v_i)\ge0.6$ \textbf{then}   
		\Statex \quad \quad$D_b\leftarrow D_b\cup\{r\}$
		
		\State \textbf{for each} $x\in D_s\cup D_b$ \textbf{do} $\Phi(x)\leftarrow\textsc{Adjudicate}(x,I)$
		
		\State \textbf{return} $\{x:\Phi(x)=\textsc{confirmed}\}$
		
	\end{algorithmic}
\end{algorithm}


%
%

\section{Evaluation}
\label{sec:eval}

%
%
%
%
%

\subsection{Experimental Setup and Ethics}


\label{subsec:setup}
\textbf{Experimental setup and MCPGuard implementation.}
We analyze Python SDK 1.26.0 and TypeScript SDK 1.29.0 against revision 2025-11-25 of the MCP specification~\cite{mcp-python,mcp-typescript,mcp-spec}, as these two SDKs had the highest number of downloads~\cite{mcp-aaif}. We freeze each SDK environment and run all PoCs against it.
Experimental evaluation was completed by July 10.
In method implementation, MCPGuard consists of a language-independent analysis core and SDK adapters to support dynamic adaptation to different SDKs. The Python adapter reflects over Pydantic models, whereas the TypeScript adapter uses a Node.js bridge to inspect the compiled Zod schemas. Both adapters produce the same normalized model representation, allowing them to reuse the structural comparison, behavioral audit, and verdict logic. The behavioral audit uses a hosted GPT-5.5 with three votes per located rule and a 4,000-character slice cap.
For dynamic validation, MCPGuard submits constraint-specific witnesses to the actual Pydantic constructors or Zod \texttt{safeParse}. Confirmed findings are emitted as independently executable \texttt{pytest} or \texttt{vitest} PoCs.
The behavioral audit uses a hosted GPT-5.5, three votes per located rule, and a 4,000-character slice cap. 

\addtocounter{table}{1} 
\begin{table*}
	\caption{Aggregate distribution of confirmed conformance-gap findings.}
	\label{tab:gapcats}
	\centering
	\begin{tabular*}{0.95\textwidth}{@{\extracolsep{\fill}}lllcccc@{}}
		\toprule
		\textbf{Track} & \textbf{Gap category} & \textbf{Abstract enforcement omission} & \textbf{Python} & \textbf{TypeScript} & \textbf{Cross-SDK} & \textbf{Total} \\
		\midrule
		Structural & Requiredness & Required-on-wire constraint is not enforced & 1 & 2 & 1 & 3 \\
		& Numeric range & Bound validator is absent & 3 & 0 & 0 & 3 \\
		& URI format & URI syntax validator is absent & 3 & 12 & 2 & 15 \\
		& URI-template format & URI-template syntax validator is absent & 2 & 2 & 2 & 4 \\
		& Byte encoding & Encoding-format validator is absent & 2 & 0 & 0 & 2 \\
		\cmidrule(lr){1-7}
		\multicolumn{3}{@{}l}{\textbf{Structural subtotal}} & \textbf{11} & \textbf{16} & \textbf{5} & \textbf{27} \\
		\midrule
		Behavioral & Protocol action & A mandatory validation before action is absent & 1 & 1 & 1 & 2 \\
		& Capability negotiation & Advertised capability is not fully enforced & 1 & 1 & 1 & 2 \\
		& Authorization policy & A mandatory authorization constraint is not enforced & 0 & 1 & 0 & 1 \\
		\cmidrule(lr){1-7}
		\multicolumn{3}{@{}l}{\textbf{Behavioral subtotal}} & \textbf{2} & \textbf{3} & \textbf{2} & \textbf{5} \\
		\midrule
		\multicolumn{3}{@{}l}{\textbf{Total}} & \textbf{13} & \textbf{19} & \textbf{7} & \textbf{32} \\
		\bottomrule
	\end{tabular*}
\end{table*}

\addtocounter{table}{-2} 
\begin{table}[t]
	\caption{Specification-to-SDK analysis coverage. 
Numerators count the number of specification units for which MCPGuard has detected the corresponding SDK modeling.}
	\label{tab:coverage}
	\centering
	\footnotesize
	\begin{tabular}{@{}lcc@{}}
		\toprule
		\textbf{Analysis unit} & \textbf{Python} & \textbf{TypeScript} \\
		\midrule
		Structural type & 108/120 (90.0\%) & 115/120 (95.8\%) \\
		Structural field & 227/254 (89.4\%) & 245/254 (96.5\%) \\
		Behavioral rule & 349/485 (72.0\%) & 339/485 (69.9\%) \\
		\bottomrule
	\end{tabular}
\end{table}
\addtocounter{table}{1} 

\textbf{Baselines.}
We compare MCPGuard with two publicly available MCP-specific conformance tools: the official MCP conformance testing framework v0.2.0-alpha.9 (official suite) and the Yaw Labs MCP Compliance suite v0.16.3 (Yaw suite)~\cite{mcp-conformance,yaw-compliance}.
The official framework represents the protocol project's reference testing practice and evaluates client or server implementations through scenario-based protocol interactions. The Yaw suite provides an independent, capability-aware assessment covering transport, lifecycle, MCP features, error handling, and schema-related checks, and reports required and optional results separately. Together, they represent official and third-party approaches to black-box MCP conformance testing.
Notably, baselines test specific SDK implementation endpoints (\ie client or server node) and evaluate observable endpoint behavior, while MCPGuard checks the execution of the underlying SDK internals and is unrelated to the implementation endpoints.

\textbf{Ethics and disclosure.}
We have reported the gap findings privately to SDK developers and received their reproduction and confirmation. We will disclose the discovered gaps once the developers have fixed them.
In this paper, we omit exploit-oriented details to prevent malicious use. 
Remote probing of open MCP servers uses only non-offensive fields. 
Each probe sent one malformed message and caused no network attack, such as denial of service, flooding, or resource exhaustion.

In our experiments, we aim to answer: 
\begin{itemize}
    \item \textbf{RQ1 (\S~4.2):} How much of the MCP specification does MCPGuard cover, and how effective is it at identifying conformance gaps?
    \item \textbf{RQ2 (\S~4.3):} Do SDK-level conformance gaps propagate systematically to downstream MCP applications?
    \item \textbf{RQ3 (\S~4.4):} How does MCPGuard compare with existing baselines in detecting conformance gaps?
    \item \textbf{RQ4 (\S~4.5):} How do the key components of MCPGuard contribute to its effectiveness and efficiency?
\end{itemize}

\subsection{RQ1: Effectiveness}
\label{subsec:rq1}

\textbf{Specification-to-SDK coverage.}
We measure the specification coverage of MCPGuard at three levels. 
At the type level, JSON Schema consists of 120 named object types, and a type is covered when MCPGuard aligns it with an SDK model. 
At the field level, we measure 254 explicit constraint instances attached to schema fields. A field is covered when MCPGuard aligns it with an SDK parameter.
At the behavioral level, we measure whether the 485 behavioral rules extracted from spec documents can be associated with a non-empty implementation slice for subsequent auditing. Table~\ref{tab:coverage} summarizes the results.
MCPGuard covers more than 89\% of the type-level and field-level structural constraints in both SDKs. Its behavioral coverage is lower. 
The possible reason is that some rules may not be implemented in the SDKs.
Coverage measures analyzability rather than conformance; an uncovered unit is not automatically considered a conformance gap.

\textbf{Conformance gap findings.} Across both tracks, MCPGuard confirms 13 gaps in the Python SDK and 19 gaps in the TypeScript SDK. All 27 structural findings are reproduced by generated regression PoCs, and the five behavioral findings survive full-context source review and targeted protocol probes where available. The reported findings were also reproduced and confirmed by the SDK developers.
Table~\ref{tab:gapcats} groups these confirmed gaps, and reveals three characteristics of them. 
First, structural findings are concentrated in validation semantics that can disappear when rich schema types are mapped to generic SDK representations. Format-related omissions account for 21 of the 27 structural records (77.8\%): 15 concern URI syntax, four concern URI-template syntax, and two concern byte encoding. Nevertheless, the three requiredness and three numeric-range findings show that specification drift also affects whether a member must appear on the wire and which values an otherwise well-typed number may take. 
Second, the cross-language comparison reveals asymmetric conformance. 6 gaps are Python-only because TypeScript enforces the corresponding bounds or byte formats, whereas 11 are TypeScript-only because Python uses URL-aware fields for those URI members.
These contrasts indicate that their SDK modeling frameworks preserve different subsets of the same protocol contract. At the same time, 5 structural constraints recur across both SDKs, suggesting that part of the gap surface reflects a shared difficulty in translating specification semantics into run-time validators rather than isolated implementation mistakes.
Third, the behavioral findings extend beyond field-local validation. 4 of the 5 records concern guards whose correctness depends on the surrounding protocol action or negotiated capability state, while the remaining record concerns authorization-policy conformance. These findings cannot be expressed as requiredness, range, or format mismatches, which demonstrates the complementary role of the behavioral track. 

\begin{figure*}[t]
	\centering
	\includegraphics[width=\textwidth]{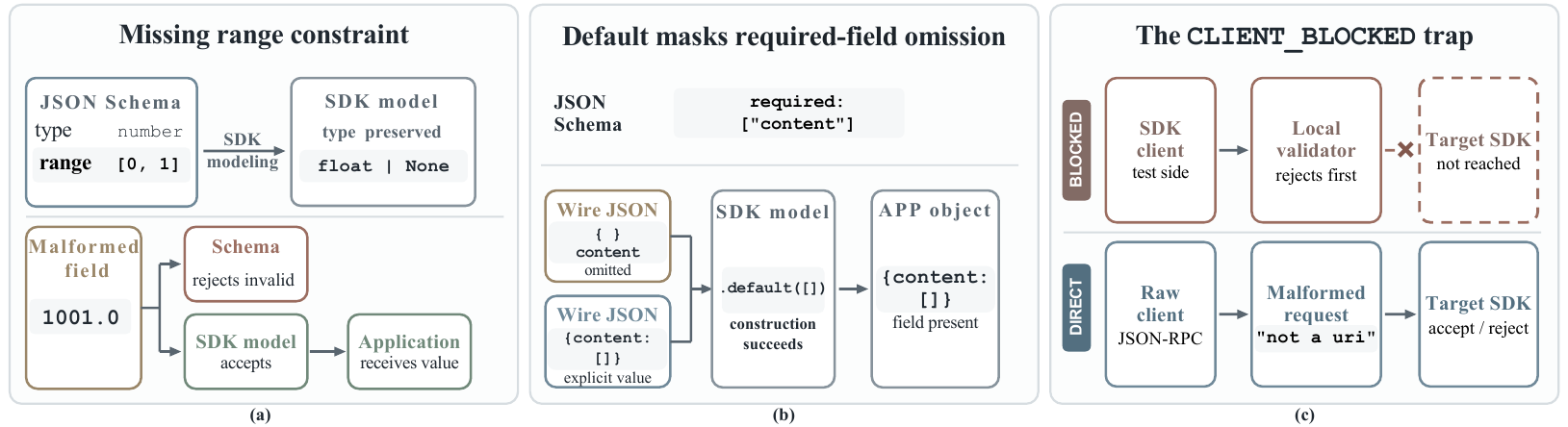}
	\caption{Three omitted conformance gap cases. (a) A bound required in JSON Schema disappears from the SDK model, so an out-of-range witness reaches application code. (b) A default makes a required wire field optional and hides the omission in the parsed object. (c) An SDK client can reject malformed input before transmission (\textsc{client-blocked}); therefore, a raw client is needed to observe target-side conformance.}
	\label{fig:abnormal-cases}
\end{figure*}

Fig.~\ref{fig:abnormal-cases}(a)--(b) illustrates two de-identified mechanisms through which structural constraints are lost during SDK modeling. In Fig.~\ref{fig:abnormal-cases} (a), the schema constrains a numeric field to a declared interval, whereas the corresponding SDK model preserves only its primitive numeric type. Consequently, an out-of-range value still passes model construction and reaches applications, even though it is invalid under the MCP JSON schema.
Fig.~\ref{fig:abnormal-cases} (b) illustrates a requiredness gap caused by an SDK default. The schema requires the member to be explicitly present in the received JSON message, but the SDK assigns it a default value when it is omitted. Model construction therefore succeeds, and the resulting object is indistinguishable from one produced from a message that explicitly supplied the default value. The SDK thus guarantees post-parse presence while failing to enforce wire-level requiredness, masking the protocol violation from downstream application code.

\subsection{RQ2: Systematicity}
\label{subsec:rq2}
\begin{table}[t]
	\caption{Results of probe testing on 32 open MCP servers.}
	\label{tab:remote}
	\centering
	\footnotesize
	\begin{tabular}{@{}lcccc@{}}
		\toprule
		\textbf{Field/path} & \textbf{Accept} & \textbf{Reject} & \textbf{Lookup} & \textbf{N.A.} \\
		\midrule
		\texttt{websiteUrl} & 32 & 0 & 0 & 0 \\
		\texttt{icon.src} & 32 & 0 & 0 & 0 \\
		\texttt{resources/read} & 1 & 15 & 5 & 11 \\
		\bottomrule
	\end{tabular}
\end{table}

We evaluate whether the confirmed gaps exist in open MCP servers to verify the systematicity of these gaps.
Specifically, we probe 32 MCP servers open on the Internet, which are provided by Adobe, Google, Yahoo, Microsoft \etc~\cite{mcp-servers}. 
These servers are built using the TypeScript language, so we attempt to test them by exploiting gaps discovered in TypeScript.
As Fig.~\ref{fig:abnormal-cases} (c) shows, the probe uses an SDK-agnostic raw HTTP client because an SDK-side client-blocked result says nothing about target conformance.
Then, we use the client to handshake with the servers and send three types of invalid fields to them, including websiteUrl, icon.src, and resources/read. 
The former two probes target handshake metadata, while the third supplies a malformed resource identifier to an active resource request. 
Each server receives only one request per probe.

These servers' responses are shown in Table~\ref{tab:remote}. All 32 servers that complete initialization accept both malformed handshake metadata values, demonstrating the systematicity of the two gaps.
The active resource request exhibits different results. Of the 32 servers, ten do not expose the relevant capability, and one reports the method as unavailable. Among the remaining 21 servers, 15 explicitly reject the malformed identifier (71.4\%), five continue to resource lookup and return a not-found error (23.8\%), and one accepts the request (4.8\%). Lookup and accept errors show that the malformed value reaches the application-level resource handling. 
These results also reveal a clear path-dependent difference.
The shared URI-format gaps are consistently reachable and systematic through handshake metadata paths, while deployment-specific checks can partially mitigate them on active request paths.
The agreement between the SDK-level findings and the deployment observations provides evidence that SDK-level conformance gaps can propagate systematically into downstream applications.

\subsection{RQ3: Comparison with Baselines}
\label{subsec:rq3}
%

\begin{table}[t]
    \caption{Mandatory conformance results on the same SDK versions. }
    \label{tab:baselines}
    \centering
    \footnotesize
    \begin{tabular}{@{}llcc@{}}
        \toprule
        \textbf{Tool} & \textbf{Target} & \textbf{Python SDK} & \textbf{TypeScript SDK} \\
        \midrule
        Official suite & endpoint behavior
            & 0
            & 0  \\
        Yaw suite & endpoint behavior
            & 0
            & 0 \\
        MCPGuard & SDK enforcement
            & 13 confirmed 
            & 19 confirmed \\
        \bottomrule
    \end{tabular}
\end{table}

We further compared MCPGuard with the official suite and Yaw suite. Both baselines drive the SDK reference servers over streamable HTTP. The official suite executes 30 applicable scenarios and performs 42 applicable checks for each SDK. The Yaw suite performs 19 required checks for the Python SDK and 20 required checks for the TypeScript SDK. 

Table~\ref{tab:baselines} reports their check results. The official suite passes all 42 applicable checks for both SDKs and reports no conformance gap. The Yaw suite also passes all required checks and reports no required-item failure. In contrast, MCPGuard confirms 13 gaps in the Python SDK and 19 in the TypeScript SDK. Neither baseline exposes any of the 32 MCPGuard findings.
The difference follows from the evaluation target and oracle. Both baselines are black-box and endpoint-oriented. They drive a running reference server over Streamable HTTP and check observable properties such as lifecycle transitions, response envelopes, status codes, and supported features. Their mandatory scenarios mainly use protocol-conforming inputs. 
Consequently, these baselines can expose the gaps only if they deliberately send the relevant malformed value, and a client built on the same SDK may block the value before transmission. 
In comparison, MCPGuard examines a different level of conformance. It compares specification constraints directly with the shared SDK models and logic, and then constructs negative witnesses for the constraints that appear unenforced. Therefore, it can determine whether the SDK rejects malformed values. 

Notably, the results do not mean that black-box conformance testing is unnecessary. The baselines verify lifecycle, transport, and endpoint behavior, whereas MCPGuard examines the processing of negative inputs and behaviors at the SDK boundary. The methods are complementary, and a complete regression strategy can combine endpoint-oriented conformance tests with specification-to-SDK validation.

\subsection{RQ4: Component Contribution}
\label{subsec:rq4}

We conduct ablation studies on the components to verify their contributions. These tested components include the dual-track mechanism, code slicing, and dynamic adjudicator.

\textbf{Track contribution.}
Removing either track leaves a different blind spot for conformance testing, resulting in fewer detected gaps. A prose-only audit does not provide a scalable inventory of hundreds of field-level validators, while a schema-only diff cannot express session conditions (such as which elicitation mode was negotiated). 
The confirmed structural and behavioral gaps have no shared keys, indicating that their mechanisms are complementary.

\begin{table}[t]
    \caption{Reduction of the behavioral audit context through rule-guided slicing.}
    \label{tab:slicing}
    \centering
    \footnotesize
    \begin{tabular}{@{}lrrrr@{}}
        \toprule
        \textbf{SDK} & \textbf{All} & \textbf{Touched} & \textbf{Proportion} & \textbf{Reduction} \\
        \midrule
        Python & 606 & 115 & 19.0\% & 81.0\% \\
        TypeScript & 1,059 & 91 & 8.6\% & 91.4\% \\
        \bottomrule
    \end{tabular}
\end{table}

\textbf{Slicing contribution.}
We evaluate whether the slicing mechanism reduces the context submitted to the LLM. For each SDK, we count the unique functions touched by at least one behavioral slice and compare them with all functions identified in the SDK. As shown in Table~\ref{tab:slicing}, the slices touch only 115 of 606 Python functions (19.0\%) and 91 of 1,059 TypeScript functions (8.6\%). 
Many remaining functions implement serialization, transport, tool registration, and other auxiliary operations that do not directly enforce the selected normative rules. Consequently, slicing concentrates the behavioral audit on likely function paths and avoids supplying most unrelated implementation code to the LLM.
This reduction does not indicate lower specification coverage. MCPGuard still obtains implementation slices for 349 rules in Python and 339 rules in TypeScript, as reported in Table~\ref{tab:coverage}. Instead, the two measurements show that MCPGuard covers a large fraction of the behavioral rules while exposing the LLM to only a small set of the SDK functions.


\begin{figure}[tbp]
	\centering
	\includegraphics[width=0.8\columnwidth]{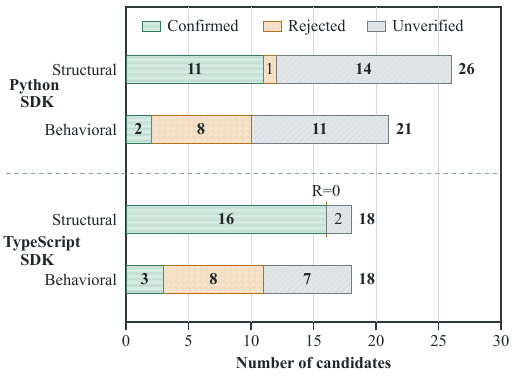}
	\caption{The adjudication results of candidate gaps.}
	\label{fig:adjudication}
\end{figure}

\textbf{Contribution of dynamic adjudication.}
Fig.~\ref{fig:adjudication} quantifies how the adjudicator separates reportable findings from the candidates generated by the two analysis tracks. For the Python SDK, it confirms 11 of 26 structural candidates and 2 of 21 behavioral candidates; for the TypeScript SDK, it confirms 16 of 18 structural candidates and 3 of 18 behavioral candidates. Aggregated across both SDKs, 27 of 44 structural candidates (61.4\%) and only 5 of 39 behavioral candidates (12.8\%) survive adjudication.
This difference reflects the distinct roles of the two tracks. Structural candidates originate from deterministic schema-to-validator mismatches and can usually be checked by constructing a kind-specific witness against the actual SDK validator. Behavioral auditing relies on an LLM to judge a rule against a relevant code slice, which may omit potential checks implemented in callers, callees, session layers, or other protocol states. The adjudicator therefore reopens the complete execution context and constructs a stateful PoC that reproduces the suspected violation.
The removal of 51 unsupported candidates demonstrates that static mismatches and LLM judgments are useful for candidate generation but are insufficient as conformance findings, while the dynamic adjudicator enables accurate detection of real gaps.

\section{Conclusion}

This paper presented MCPGuard, a specification-to-SDK analysis framework for detecting conformance gaps in MCP SDKs. 
It deterministically compares schema constraints with normalized SDK validators, while using rule-guided function slicing and LLM audit to identify potentially missing behavioral guards. An evidence-aware adjudicator then examines the complete source context and builds PoCs to verify the authenticity of gaps.
Applied to the official Python and TypeScript SDKs, MCPGuard confirms 13 and 19 conformance gaps, respectively, whereas neither of the two black-box baselines exposes these findings. Moreover, test results on open MCP servers show that those gaps can propagate systematically into downstream applications. 
MCPGuard complements conventional suites by making the SDK conformance of specification constraints directly testable, producing actionable evidence for SDK maintainers.


\bibliographystyle{IEEEtran}
\bibliography{IEEEabrv, ref}

\end{document}